\documentclass{article} % For LaTeX2e
\usepackage{iclr2026_re-align_workshop,times}
\renewcommand{\headrulewidth}{0pt} %%

\usepackage{amsmath,amsfonts,bm}

\def\eqref#1{equation~\ref{#1}}
\def\1{\bm{1}}

\DeclareMathAlphabet{\mathsfit}{\encodingdefault}{\sfdefault}{m}{sl}
\SetMathAlphabet{\mathsfit}{bold}{\encodingdefault}{\sfdefault}{bx}{n}

\usepackage{hyperref}
\usepackage{url}
\usepackage{graphicx}
\usepackage{listings}
\title{Psychological Concept Neurons: Can Neural Control Bias Probing and Shift Generation in LLMs?}

\author{Yuto Harada \& Hiro Taiyo Hamada \\
Araya Inc.\\
\texttt{\{harada\_yuto, hamada\_h\}@araya.org} \\
\\
}

\iclrfinalcopy % Uncomment for camera-ready version, but NOT for submission.
\begin{document}

\maketitle

\begin{abstract}
Using psychological constructs such as the Big Five, large language models (LLMs) can imitate specific personality profiles and predict a user's personality. While LLMs can exhibit behaviors consistent with these constructs, it remains unclear where and how they are represented inside the model and how they relate to generation. 
To address this gap, we focus on questionnaire-operationalized Big Five concepts, analyze the formation and localization of their internal representations, and further use interventions to examine how these internal representations relate to behavioral outputs.
In our experiment, we first use probing to examine where Big Five information emerges across model depth. We then identify neurons that respond selectively to each Big Five concept and test whether enhancing or suppressing their activations can bias latent representations and label generation in intended directions. We find that Big Five information becomes rapidly decodable in early layers and remains detectable through the final layers, while concept-selective neurons are most prevalent in mid layers and exhibit limited overlap across domains.
Interventions on these neurons consistently shift probe readouts toward targeted concepts, with targeted success rates exceeding 0.8 for some concepts, indicating that the model’s internal separation of Big Five personality traits can be causally steered. At the label-generation level, the same interventions often bias generated label distributions in the intended directions, but the effects are weaker, more concept-dependent, and often accompanied by cross-trait spillover, indicating that comparable control over generated labels is difficult even with interventions on a large fraction of concept-selective neurons. Overall, our findings reveal a gap between representational control and behavioral control in LLMs: concept-selective neuron sets provide a strong handle for steering personality-related representations, but comparable control over generated labels remains substantially more limited.

\end{abstract}

\section{Introduction}
As large language models (LLMs) achieve advanced language understanding, there is growing attention on more human-like aspects such as conversational style, value judgments, and interpersonal tendencies, as well as their consistency.
In psychology, such properties have traditionally been measured using psychological constructs, which provide structured and interpretable dimensions for describing stable patterns of behavior and preference. Among these, personality offers a compact and widely adopted framework for characterizing consistent tendencies across situations. Such a framework, operationalized through questionnaires such as the Big Five Inventory 2 (BFI-2; \citet{soto2017next}), has long served as a standard tool for measuring personality in humans.

Recent evidence suggests that these frameworks can be applied to LLMs as well as humans, and research using psychological concepts and their corresponding questionnaires with LLMs is increasing.
Previous works showed that LLM outputs can be conditioned to imitate the personality of specific individuals or profiles \citep{argyle2023out, park2023generative}.
Other work infers human personality traits from dialogue or text \citep{peters2024large, ravenda-etal-2025-llms}.
These findings demonstrate that LLMs can exhibit behaviors aligned with specific personalities and understand such concepts to evaluate human text.
 
Despite this progress, most previous work remains at the level of behavioral observations, leaving it unclear how LLMs encode psychological concepts internally and how such representations relate to controllable behavioral outputs.
To address this gap, we investigate the following research questions:
\begin{enumerate}
  \item Where do psychological concepts emerge in LLMs?
  \item How are psychological concepts represented by neurons in LLMs?
  \item Can psychological concepts be causally manipulated in LLMs?
\end{enumerate}
In our experiment, we study how Big Five personality concepts defined by a questionnaire are represented and controllable inside instruction-tuned LLMs. We analyze several models of similar scale and architecture. First, we perform layer-wise probing to track where trait domain labels become linearly decodable across network depth. Second, we identify psychological concept neurons by measuring unit level selectivity for each domain, extracting units that respond selectively to questionnaire based descriptions of a given trait. Third, we test causal involvement by intervening at inference time, enhancing or suppressing the activations of selected neurons to extreme quantiles and evaluating changes in both probe readouts and a domain name generation task.

Our results reveal a consistent pattern. Big Five information becomes rapidly decodable in early layers and remains detectable through later layers. This contrasts with prior reports suggesting that persona related behavioral representations emerge primarily in deeper layers. In contrast to the broad availability of decodable information, concept-selective neurons are most prevalent in mid layers and show limited overlap across domains, indicating a localized and concept specific organization. Interventions on these neurons strongly bias probe readouts toward the targeted concept, reaching shifts of up to roughly 80\% with minimal spillover in favorable settings. For generation, the same interventions often shift output distributions in the intended direction, with effects reaching up to roughly 50\%, but with greater variability across concepts and occasional cross trait spillover. 
Overall, these findings suggest that localized concepts associated with personality representations provide a meaningful control signal, while downstream generative behavior depends on additional mechanisms beyond those that dominate probing.

\section{Related Work}

\subsection{Measuring Personality Traits with Questionnaires}
The Big Five is a widely used framework for comprehensively describing personality along five trait dimensions: Extraversion, Agreeableness, Conscientiousness, Negative Emotionality, and Open-Mindedness \citep{soto2017next}. In psychology, these traits are treated as psychological constructs for explaining personality. A psychological construct refers to a latent property, such as intelligence, anxiety, or personality traits, that is not directly observable but is operationally defined and measured through observable indicators, including responses to questionnaire items.  

The Big Five Inventory 2 (BFI-2), used in this study, is a revised version of the original Big Five Inventory. In addition to the five broad domains, it hierarchically measures fifteen facets, resulting in a total of sixty questionnaire items.

\subsection{Personality Traits and Large Language Models}
In recent years, an increasing number of studies have examined personality traits in large language models using frameworks such as the Big Five. Some studies evaluate the personality tendencies of models themselves by administering personality questionnaires \citep{10.1093/pnasnexus/pgae533, sorokovikova-etal-2024-llms}. Others condition models to imitate specific individuals or predefined personality profiles \citep{argyle2023out, park2023generative}. Additional work infers human personality traits from dialogue or text \citep{peters2024large, ravenda-etal-2025-llms}, or aims to induce and align desirable personality tendencies in models \citep{jiang2023evaluating, zhu2025personality}. Collectively, these studies suggest that large language models can acquire and express psychological constructs related to personality through training and conditioning, and that such constructs can be manipulated at the behavioral level.  

At the same time, most of these approaches rely on behavioral observations, such as questionnaire responses or generated text. As a result, the internal representations underlying personality related behavior, and the specific components that contribute to generation, remain largely unclear.

In parallel, recent work has begun to localize and manipulate internal representations associated with persona or personality expressed in dialogue. For example, some studies use layer-wise probing to extract response personality and apply the resulting readout directions to edit personality at inference time \citep{ju2025probing}. Other work analyzes at which layers persona representations become separated and examines their localization within the model \citep{cintas2025localizing}. While these studies are important in demonstrating the controllability of persona as a behavioral property, our work differs in that we focus not on behavior itself, but on how psychological constructs are recognized and internally represented by the model.

\begin{figure}[t]
    \centering
    \includegraphics[width=1.0\linewidth]{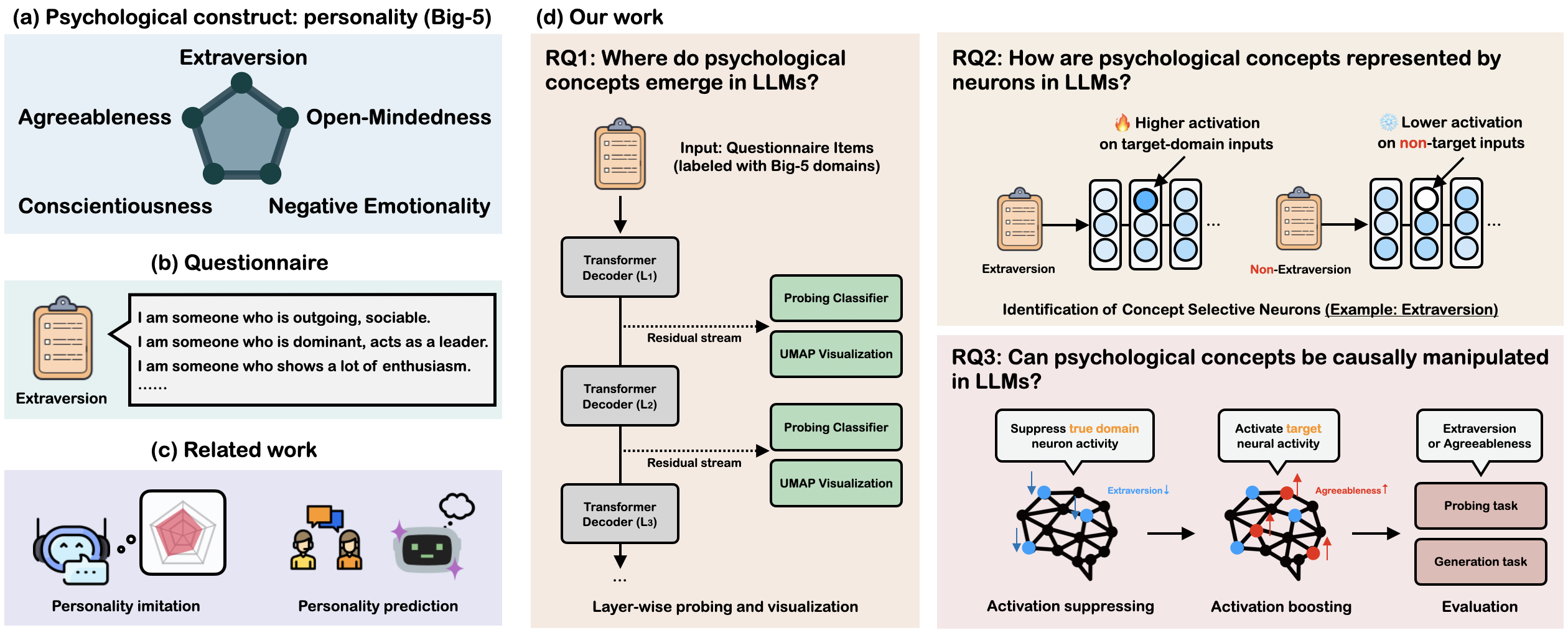}
    \caption{
    Overview of the background and experimental design.
    \textbf{(a)} The Big Five framework defines human personality using five latent trait domains, which are treated in psychology as psychological constructs (i.e., abstract properties that are not directly observable but operationally defined).
    \textbf{(b)} Example item from the Big Five Inventory-2 (BFI-2), a standard personality questionnaire, for Extraversion. Respondents rate how well the statement describes them on a 5-point Likert scale.
    \textbf{(c)} Related trends. Recent work increasingly applies psychometric personality concepts to LLMs, showing that models can exhibit and be conditioned toward stable personality-like behaviors. However, these findings are largely behavioral, leaving the underlying internal representations and mechanisms unclear.
    \textbf{(d)} Our approach is organized around three questions: RQ1 where questionnaire-defined concepts emerge across layers, RQ2 which neurons are selective for each concept, and RQ3 whether intervening on those neurons can causally steer readouts and label generation.
    }
    \label{fig:overview}
\end{figure}

\section{Methods}
\label{sec:methods}
Figure \ref{fig:overview} provides an overview of this experiment. This section describes the methodology employed in each analysis.

\subsection{Layer-wise probing}
Layer-wise probing evaluates how well concept labels can be read out from internal representations at each layer using linear classifiers \citep{alain2017understanding}.  
Let $x$ denote an input sentence and $y \in \{1, \dots, K\}$ its corresponding label, where $K$ is the number of target labels. When using Big Five domain labels, we have $K = 5$. For an input $x$, the model produces token representations $\mathbf{h}_{\ell,t}(x) \in \mathbb{R}^d$ at each layer $\ell$. In this work, we use the representation of the final token as the sentence representation and denote it by $\mathbf{h}_{\ell}(x)$.
For each layer $\ell$, we train a linear classifier
\begin{equation}
p_{\ell}(y \mid x) = \mathrm{softmax}(\mathbf{W}_{\ell} \mathbf{h}_{\ell}(x) + \mathbf{b}_{\ell})
\end{equation}
and compare classification performance across layers to examine where conceptual information becomes easier to read out.

\subsection{Identification of concept-selective neurons}
Next, we identify neurons that activate selectively for each concept. Following the approach of Suau et al.\ \citep{suau2020findingexpertstransformermodels}, we define concept-selectivity by ranking input sentences according to unit activations and evaluating how well this ranking separates concept labels using Average Precision.
Let $z_{\ell,j}(x)$ denote the pre-activation of unit $j$ in the MLP at layer $\ell$, evaluated at the final token position for $x$. In this work, we represent unit $j$’s response to sentence $x$ by this value and define
\begin{equation}
a_{\ell,j}(x) = z_{\ell,j}(x).
\end{equation}
Given a dataset $\{(x_i, y_i)\}_{i=1}^N$, we define positive examples for concept $c$ as inputs satisfying $y_i = c$, and negative examples as all other inputs. Each unit $j$ is treated as a scoring function that ranks inputs in descending order of $a_{\ell,j}(x_i)$. Units that show high activations for inputs of concept $c$ and low activations for other inputs will rank positive examples higher, resulting in a higher Average Precision score.
The selectivity of unit $j$ for concept $c$ is computed as the Average Precision of this ranking. For each concept $c$, units with high Average Precision are extracted as concept-selective neurons. In practice, we use the set of top ranked units $S_{\ell}^{(c)}$ for each concept.
Details such as the number of selected units are described in the experimental section.

\subsection{Activation boosting and suppressing}
To test whether the extracted concept-selective neurons are causally involved in model outputs, we intervene on their activations at inference time. Our generation task requires the model to output the corresponding domain name as a single token given an input sentence, and interventions are applied only at the first generation step.
We perform interventions by setting unit activations to quantile values. First, for each unit, we compute quantiles over unit responses across the entire dataset,
\begin{equation}
q_{\ell,j}^{(p)} = \mathrm{Quantile}_p(\mathbf{s}_{\ell,j}),
\end{equation}
where $p \in \{0.01, 0.99\}$. For the set corresponding to the target concept $c^+$, activations are replaced with the upper quantile, while for the set corresponding to the suppressed concept $c^-$, activations are replaced with the lower quantile.
We then observe how this intervention changes probing classification performance and the output distribution of the generation task, in order to assess the causal contribution of concept-selective neurons.

\section{Experimental Setup}

\paragraph{Data}
We use questionnaire items from the Big Five Inventory~2 (BFI-2) as text definitions of personality-related psychological constructs. The dataset contains 60 short statements, each labeled with one of the five Big Five domains. Throughout the paper, we treat domain prediction from an item as a 5-way classification problem.

\paragraph{Prompting}
For probing and representation analyses, we prepend a fixed classification instruction to each item and ask the model to output exactly one label from a closed set (domain names). We use the same prompt template across models and tasks; the full template is provided in Appendix~\ref{app:prompts}.

\paragraph{Models}
We run experiments on a set of instruction-tuned pretrained LLMs. Our primary model is \texttt{Meta-Llama-3-8B-Instruct} \citep{dubey2024llama}, which we use for analyses that require detailed inspection (e.g., unit-level selectivity and interventions). To confirm that key trends are not model-specific, we additionally evaluate \texttt{google/gemma-7b-it} \citep{gemmateam2024gemmaopenmodelsbased}, \texttt{Qwen/Qwen2.5-7B-Instruct} \citep{qwen2025qwen25technicalreport}, and \texttt{mistralai/Mistral-7B-Instruct-v0.1} \citep{jiang2023mistral7b}. For model-size comparisons within families, we also include \texttt{google/gemma-2b-it} \citep{gemmateam2024gemmaopenmodelsbased}, \texttt{Qwen/Qwen2.5-3B-Instruct} \citep{qwen2025qwen25technicalreport}, and \texttt{meta-llama/Llama-3.2-3B-Instruct} \footnote{https://huggingface.co/meta-llama/Llama-3.2-3B-Instruct}.

\paragraph{Representations}
For layer-wise probing and visualization, we use residual stream representations at each layer. To obtain the overall meaning representation of the sentence, we use the residual stream of the final token (i.e., the hidden state of the final token) as input. For concept-selective neurons, we analyze MLP units using their pre-activation values (as described in Section~\ref{sec:methods}).

\paragraph{Probing}
Layer-wise probing is performed with linear classifiers and evaluated using leave-one-out cross validation (LOOCV). We adopt a linear probe to assess whether domain information is linearly decodable at each layer, rather than to maximize predictive performance with a complex model. Because the dataset consists of only 60 questionnaire items, LOOCV allows us to use nearly all samples for training while still obtaining an out-of-sample estimate for every item. We report the average classification accuracy over all held-out items.

\paragraph{Interventions}
For probe interventions, we intervene at the layer with the highest probing accuracy. We evaluate interventions only on inputs that are correctly classified at baseline, in order to measure intervention-induced transitions rather than corrections of probe errors.
For each selected unit, the activation boost is set to the 99th percentile value calculated using positive examples in each domain, and the suppression is set to the 1st percentile value calculated using negative examples in each domain.
% For each selected unit, activation boosting sets values to the 99th percentile and suppression sets values to the 1st percentile computed over the dataset.
For intervention-based generation, we apply the intervention only at the first generation step and compare the output probabilities of label tokens under a classification-style prompt.
In this experiment, the intervened neurons are the top 30\% of the model or layer.

We quantify intervention effects using Targeted Success Rate (TSR), defined as the fraction of trials where the predicted label equals the intervention target label. We also report spillover, defined as the fraction of trials where the prediction changes to a label that is neither the original (true) label nor the target label.

\section{Results and Discussion}

\begin{figure}
    \centering
    \includegraphics[width=0.95\linewidth]{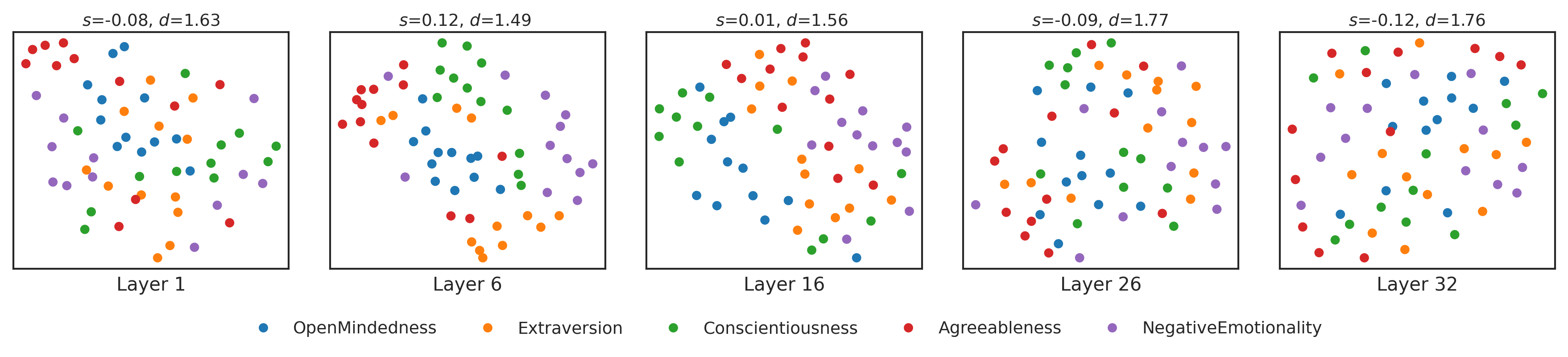}
    \caption{A two-dimensional visualisation of the residual stream used in the probing experiment. Each point corresponds to a questionnaire item, colored by its domain label, and the layer index is indicated in the panel title. \textit{S} denotes silhouette score, \textit{D} denotes intra-cluster mean distance.}
    \label{fig:visualization}
\end{figure}

\begin{figure}
    \centering
    \includegraphics[scale=0.46]{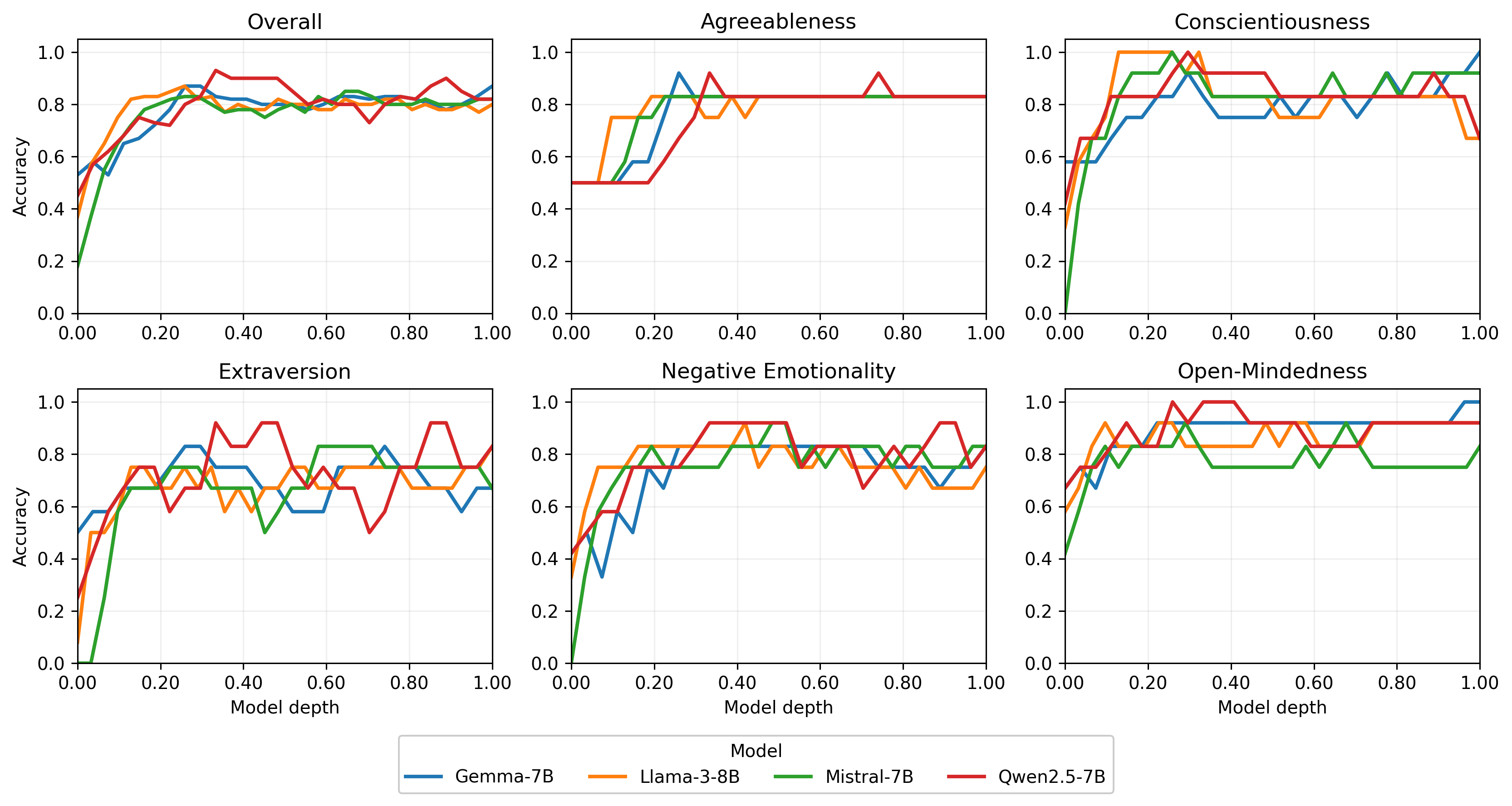}
    \caption{Results of layer-wise probing experiments. The x-axis indicates the model layer, which varies between models; hence it is shown as a proportion from the initial layer to the final layer. The results for each domain within the Big Five, with ``Overall'' representing the average result across all domains.}
    \label{fig:probing}
\end{figure}

\subsection{RQ1: Where do psychological concepts emerge in LLMs?}

\subsubsection{Visualisation of latent representations}
\label{sec:visualization}
To complement probing, we visualize residual stream representations with UMAP \citep{mcinnes2018umap} and quantify how domain-wise separation changes across layers (Figure~\ref{fig:visualization}). At shallow depth, the five domains tend to be more visually separated, whereas deeper layers show progressively stronger mixing in the embedded space. This qualitative trend is reflected by the accompanying metrics: the silhouette score $S$ is relatively higher in shallow-to-mid layers and tends to decrease in later layers, while the intra-cluster distance $D$ generally increases with depth, indicating reduced compactness.
The full layer-wise trajectories of $S$ and $D$ across the model are provided in Appendix~\ref{app:visualization_metrics}. At the same time, probing results in Section~\ref{sec:probing} show that Big Five information remains linearly decodable across depth, even as visual domain separation becomes weaker.

\subsubsection{Layer-wise probing}
\label{sec:probing}
Figure~\ref{fig:probing} reports probe accuracy across depth, where linear classifiers are trained to predict Big Five domain labels from layer representations. Each panel shows results for the five domains and their average.
Across all models, probe accuracy rises sharply in the early layers and reaches a high level by around normalized depth 0.2, after which Big Five information remains linearly decodable through the final layers. While trajectories differ by domain, the dominant pattern is an early emergence followed by a broad plateau, indicating that representations supporting trait-domain categorization are established quickly and then persist throughout the network.

\begin{figure}
    \centering
    \includegraphics[scale=0.42]{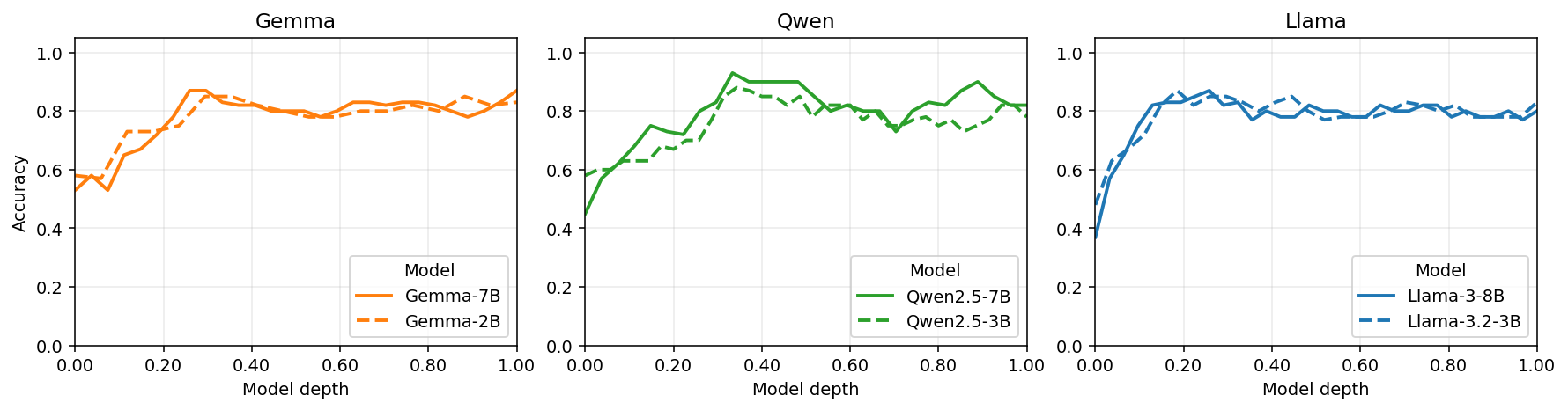}
    \caption{Results of probing experiments for each comparable model size. The dashed line indicates smaller model sizes within the same series.}
    \label{fig:model_size}
\end{figure}

We additionally compare model sizes within the same family (Figure~\ref{fig:model_size}). The depth profiles are highly similar between smaller and larger variants, suggesting that the representational structure required to separate questionnaire-defined trait domains is largely scale-robust in the 2--8B range. In other words, for the task of distinguishing Big Five domains from questionnaire items, relatively small instruction-tuned models appear to already acquire sufficient cues and organization for linear readout.
This scale-robustness contrasts with findings that certain structured representations become stronger with model size, such as geographic coordinate representations \citep{gurnee2024language}. A plausible explanation is that questionnaire-defined personality concepts arise less from memorizing specific factual entities and more from combining common vocabulary, evaluative language, and contextual patterns, which smaller models can learn stably, leading to weak size dependence in linear readout.

This early emergence contrasts with prior work that localizes persona- or response-personality-related separability in later layers. For instance, studies on persona representations report that separation becomes most pronounced toward deeper blocks \citep{cintas2025localizing}, and layer-wise analyses of response personality under Big Five style instructions similarly suggest that personality signals for responding concentrate in middle-to-upper layers \citep{ju2025probing}. We attribute the difference primarily to the target phenomenon and task setting: our probes target concept categorization of questionnaire items (trait-domain recognition), whereas persona and response-personality work emphasizes behavioral realization during generation (e.g., how an instructed persona is reflected in the produced response). Under this view, concept-level categorization cues can become available earlier in processing, while behavioral control signals tied to generating persona-consistent outputs may rely more heavily on later-layer computations.

\subsection{RQ2: How are psychological concepts represented by neurons in LLMs?}

\subsubsection{Identifying and localizing concept-selective neurons}
\begin{figure}
    \centering
    \includegraphics[scale=0.30]{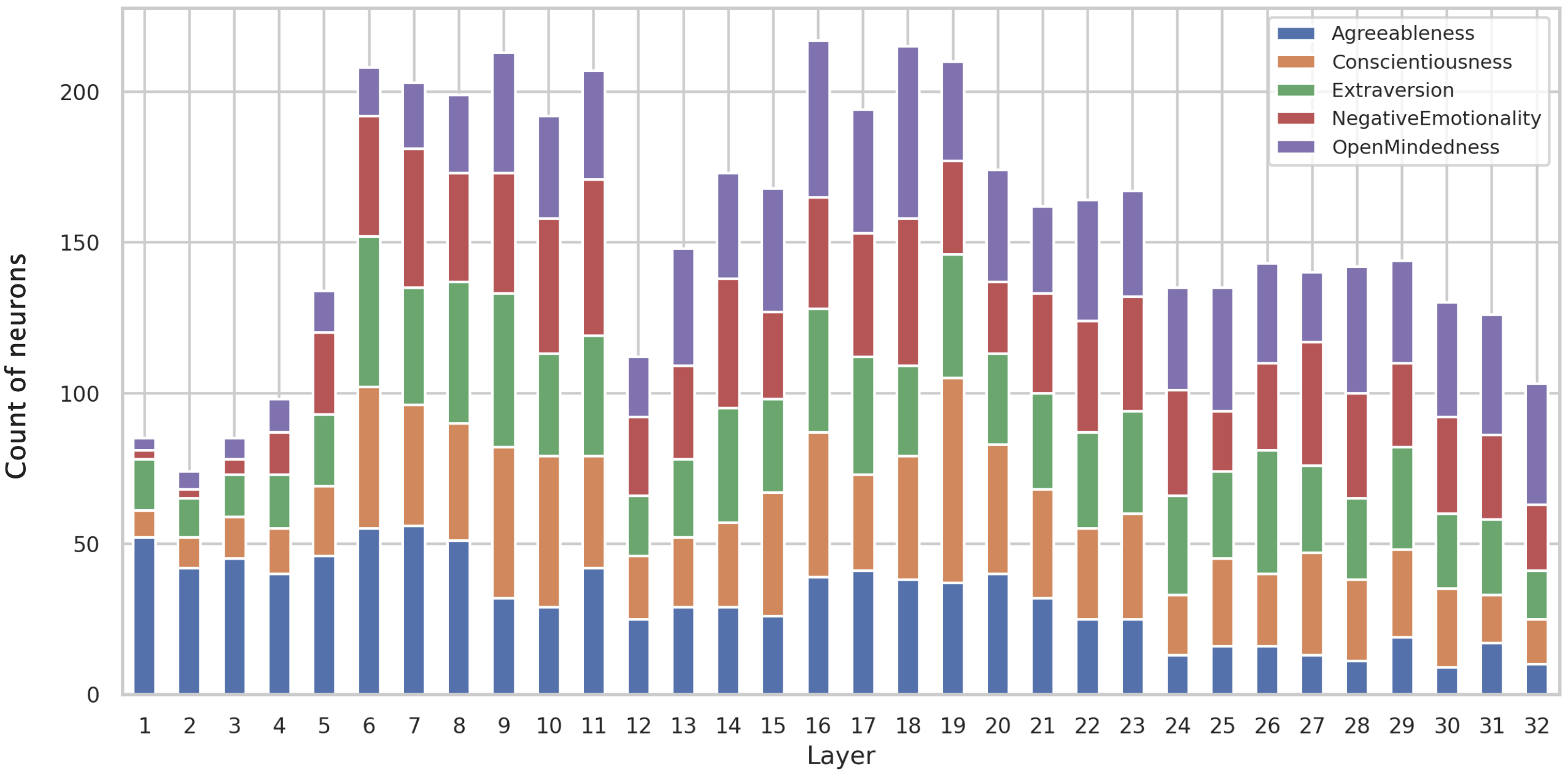}
    \caption{Layer-wise distribution of concept-selective neurons. For each domain, we select the top 1000 units across the entire model and plot their layer locations.}
    \label{fig:expert}
\end{figure}

Figure~\ref{fig:expert} shows the layer-wise distribution of concept-selective neurons extracted from Llama~3, where the top 1000 units are selected independently for each domain. The distribution varies with depth, with two clear concentration bands around layers 6–11 and 16–19. One natural interpretation is that concept-selectivity is supported by at least two functionally distinct representational regimes along depth: an earlier stage where domain-relevant cues become organized into selective units, and a later stage where these selectivities reappear or are reused in a different context of processing. In other words, concept-selective units are not confined to a single narrow layer range; instead, selectivity concentrates in multiple depth regions across the network.

Beyond this global pattern, domains differ in where selectivity appears. Agreeableness-selective units are already abundant in the earliest layers, whereas Negative Emotionality and Open-Mindedness are comparatively sparse at shallow depth and become more prevalent in mid-to-late layers. The early prevalence of Agreeableness-selective units is consistent with the incentives of instruction tuning, which rewards cooperative and socially aligned responses. More broadly, the onset of selectivity differs by domain, indicating that the depth profile is not simply the same pattern repeated across traits but reflects domain-specific emergence of concept-selectivity.

\subsubsection{Cross-domain overlap of concept-selective neurons}
Figure~\ref{fig:overlap} analyzes how much the extracted unit sets overlap across domains. We report the observed overlap among the top 10\% units as a Jaccard coefficient, normalized by the expected overlap under random selection. All domain pairs show chance-normalized overlap far below 1.0, indicating that concept-selective unit sets are largely separated rather than shared. At the same time, overlap magnitudes vary across pairs, suggesting partial sharing for specific domain combinations. This overall sparsity of overlap suggests that many concept-selective units are primarily tied to a single domain, rather than being broadly reused across traits. That said, our selection procedure is explicitly optimized for identifying units that distinguish one domain against all others; it does not directly search for multi-domain units that respond jointly to particular concept combinations (e.g., units selective for both Agreeableness and Conscientiousness). An interesting direction for future work is therefore to extend the analysis to identify such multi-concept units, which may capture shared structure across traits beyond what is reflected by one-vs-rest selectivity.

This pattern is informative when contrasted with standard psychometric structure in humans. Big Five domains are not perfectly independent and are often summarized by two higher-order metatraits: $\alpha$/$\beta$ \citep{digman1997higher} or Stability/Plasticity \citep{deyoung2006higher}, where Stability groups Agreeableness and Conscientiousness with Emotional Stability (low Negative Emotionality) and Plasticity links Extraversion with Openness/Intellect (i.e., Open-Mindedness in BFI-2 terminology). Our unit-set overlap does not simply mirror these trait-score correlations: for example, pairs linked under Stability such as Agreeableness and Conscientiousness show relatively low overlap, whereas Extraversion exhibits comparatively higher overlap with multiple domains. This highlights that psychometric covariation and mechanistic sharing among concept-selective units capture different notions of relatedness, and therefore need not align.

\begin{figure}[t]
    \centering
    \includegraphics[scale=0.38]{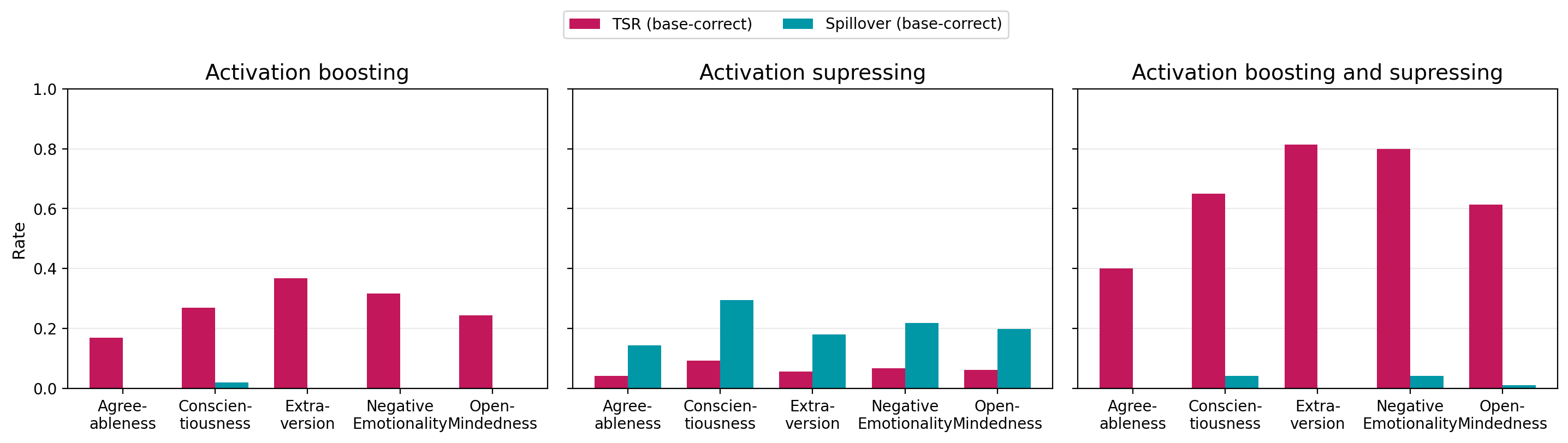}
    \caption{Results of intervention experiments on layer-wise probing. We intervened in concept-selective neurons in two ways: activation boosting involved patching to the top 99th percentile of activation values in sentences representing the target concept, while activation suppressing involved patching to the bottom 1st percentile of activation values in sentences not representing the target concept. The figure on the right shows both applied.}
    \label{fig:probing_intervention}
\end{figure}

\subsection{RQ3: Can psychological concepts be causally manipulated in LLMs?}

\subsubsection{Activation boosting and suppressing: effects on probing}
Figure~\ref{fig:probing_intervention} summarizes intervention effects on layer-wise probing. We intervene at layer~16, where probing accuracy is highest, and modify the activations of the top 30\% concept-selective MLP units for each target domain. We report the targeted success rate (TSR) and spillover rates on the subset of inputs that were correctly classified by the original non-intervened probe. Inputs misclassified at baseline are excluded so that the reported transition rates reflect changes induced by the intervention rather than corrections of prior probe errors.

The three panels separate the contributions of boosting and suppression. In the left panel, activation boosting alone increases TSR with little to no spillover across domains. This indicates that amplifying target-selective activity provides a clean control signal for the linear readout, and that enhancing concept-selective units can be critical for sharpening linear separability in the probe space.

In the middle panel, activation suppressing alone weakens evidence for the true label without explicitly steering the representation toward any particular alternative. Consistent with this intent, label changes disperse across the remaining domains and spillover dominates. TSR is correspondingly smaller and is on the order of one quarter of the total spillover, suggesting that suppression primarily removes true-domain signal rather than inducing a specific competing concept.

In the right panel, applying activation boosting and suppressing together produces the strongest and most consistent targeted shifts while keeping spillover low. The pattern is consistent with an approximately additive combination of the two effects. Suppression reduces competing influence from true-domain evidence, while boosting increases evidence for the target concept, yielding a larger net movement of the probe decision. Although our interventions are restricted to MLP units and do not test attention-based mechanisms, these results suggest that MLP activations can exert substantial causal leverage over the residual-stream representations that dominate linear readout. The high success rates also highlight a potential fragility: linear probing can read out and steer aspects of the model’s internal state relatively easily in this setting.

\subsubsection{Activation boosting and suppressing: effects on generation}
\begin{figure*}[t]
  \centering
  \begin{minipage}{0.49\textwidth}
    \centering
    \includegraphics[width=\linewidth]{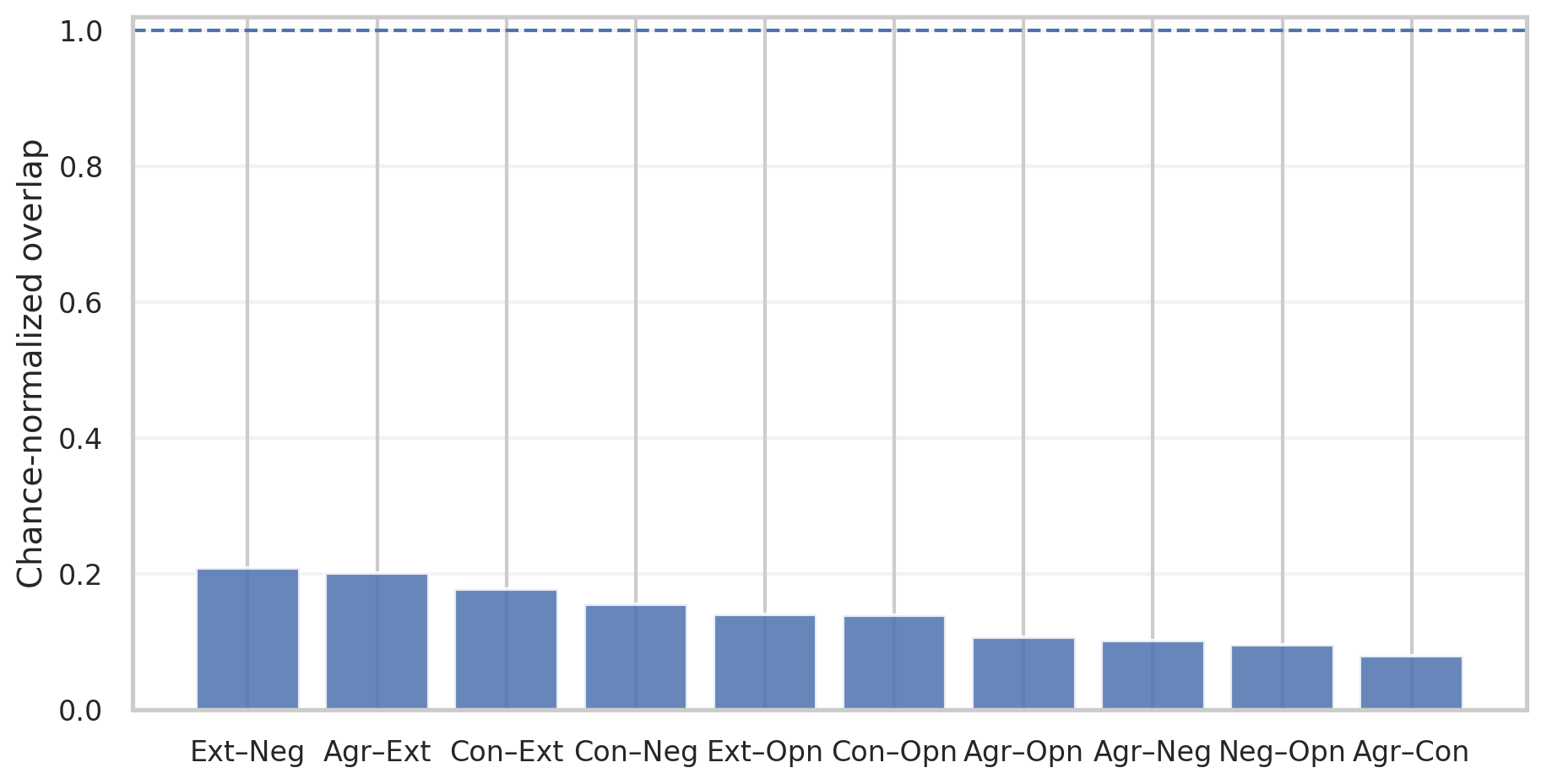}
    \caption{Analysis of overlap among concept-selective neurons. The overlap between unit sets extracted across domains is measured using the Jaccard coefficient, and the value obtained by dividing this by the random expectation is shown.}
    \label{fig:overlap}
  \end{minipage}
  \begin{minipage}{0.49\textwidth}
    \centering
    \includegraphics[width=\linewidth]{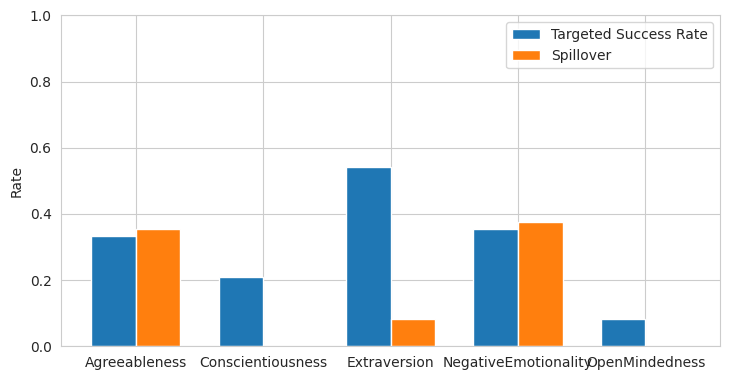}
    \caption{Results of intervention experiments on the domain name generation task.}
    \label{fig:generation_intervention}
  \end{minipage}\hfill
\end{figure*}

Figure~\ref{fig:generation_intervention} reports intervention results on the domain-name generation task. We intervene on the top 30\% concept-selective units selected across the entire model for each target domain. Compared with probing, generation shows weaker and more concept-dependent controllability. Extraversion is the most reliably inducible domain, reaching the highest targeted success with relatively small spillover. For Agreeableness and Negative Emotionality, targeted shifts are achievable, but spillover is comparably large; in many cases the model transitions to Extraversion rather than the intended domain.
This tendency aligns with the unit-set overlap analysis in Figure~\ref{fig:overlap}, where Extraversion shows comparatively higher overlap with multiple domains, suggesting that interventions may inadvertently activate shared circuitry that favors Extraversion-like outputs. Conscientiousness and Open-Mindedness exhibit smaller overall effects in generation, consistent with limited causal leverage from the selected units under this single-token prediction setting.

Even when we manipulate concept-selective units that should directly support domain discrimination, it remains difficult to fully control generation. Even though our intervention forcibly biases the activations of roughly 30\% of the selected units across the model, the model appears to maintain robustness through distributed processing in other components, which can counteract or dilute the intended bias. We also tested intervening on units from a single layer only (layer~16), but the targeted success rate was close to zero.

\section{Conclusion}
This paper shows that psychological constructs related to personality, as defined by questionnaire items, emerge in a readable form in early layers of large language models regardless of model type, and are carried by separated groups of concept-selective neurons.
Furthermore, we demonstrate that intervening on these neuron groups can systematically shift both probing outcomes and generated label distributions. Taken together, our results reveal a gap between representational control and behavioral control. Concept-level probing can be steered strongly, whereas control over generated labels is substantially more limited and more concept-dependent.
As a direction for future research, it would be valuable to extend our approach beyond personality to a wider range of psychological constructs and to test whether these interventions scale from closed-set settings to multi-token, open-ended generation. Clarifying how psychological concepts are represented in LLMs and linked to behavior would provide foundational insights toward more robust concept-level alignment between humans and LLMs.


\subsubsection*{Acknowledgments}
This work was supported by JST Moonshot R\&D Program Grant Number JPMJMS2295.

\bibliography{iclr2026_conference}
\bibliographystyle{iclr2026_conference}

\appendix

\section{Prompt Templates}
\label{app:prompts}

We use a fixed classification-style prompt for probing and representation analyses.
For each input item, the model is instructed to output \emph{exactly one} label from a closed set of domain names.
In all experiments, \texttt{\{LABELS\}} is replaced with the list of valid domain labels, and \texttt{\{ITEM\}} is replaced with the questionnaire statement.

\paragraph{Domain classification prompt.}
\begin{lstlisting}
You are a classifier. Read the statement and answer with exactly one label from the list.
Valid labels: {LABELS}. Output only the label text, nothing else.
Statement: {ITEM}

Answer with exactly one of: {LABELS}
\end{lstlisting}

\section{Supplementary Analysis for Latent Representation Visualisation}
\label{app:visualization_metrics}

This appendix provides additional details for the metrics reported alongside the UMAP visualisations in Section~\ref{sec:visualization}. We quantify the degree of domain-wise separation and compactness of the embedded representations using two scores: the silhouette score $S$ and the intra-cluster distance $D$.

\subsection{Metric definitions}
Let $\mathcal{X}=\{x_i\}_{i=1}^{N}$ be the set of samples and let each sample have a domain label $y_i \in \{1,\dots,5\}$. After computing a UMAP embedding, we obtain a 2D point $\mathbf{u}_i \in \mathbb{R}^2$ for each sample. All distances below are computed in the embedding space using Euclidean distance.

\paragraph{Silhouette score ($S$).}
For each sample $i$, we compute
\begin{itemize}
    \item $a(i)$: the average distance from $\mathbf{u}_i$ to other samples in the same label group $y_i$,
    \item $b(i)$: the minimum, over all other label groups, of the average distance from $\mathbf{u}_i$ to samples in that group.
\end{itemize}
The silhouette value for sample $i$ is then
\begin{equation}
s(i) = \frac{b(i) - a(i)}{\max\{a(i), b(i)\}},
\end{equation}
which takes values in $[-1, 1]$, where larger values indicate clearer separation from other label groups. We report $S$ as the mean of $s(i)$ over all samples.

\paragraph{Intra-cluster distance ($D$).}
For each label group $c$, let $\mathcal{I}_c=\{i \mid y_i=c\}$ be the set of indices belonging to that group. We compute the average pairwise distance within each group and then average across groups:
\begin{equation}
D = \frac{1}{5}\sum_{c=1}^{5} \frac{2}{|\mathcal{I}_c|(|\mathcal{I}_c|-1)} \sum_{\substack{i,j \in \mathcal{I}_c \\ i<j}} \lVert \mathbf{u}_i - \mathbf{u}_j \rVert_2.
\end{equation}
Smaller $D$ indicates that points within each domain group are more tightly clustered in the embedding.

\subsection{Layer-wise trends across the model}
\begin{figure}[t]
    \centering
    \includegraphics[width=0.98\linewidth]{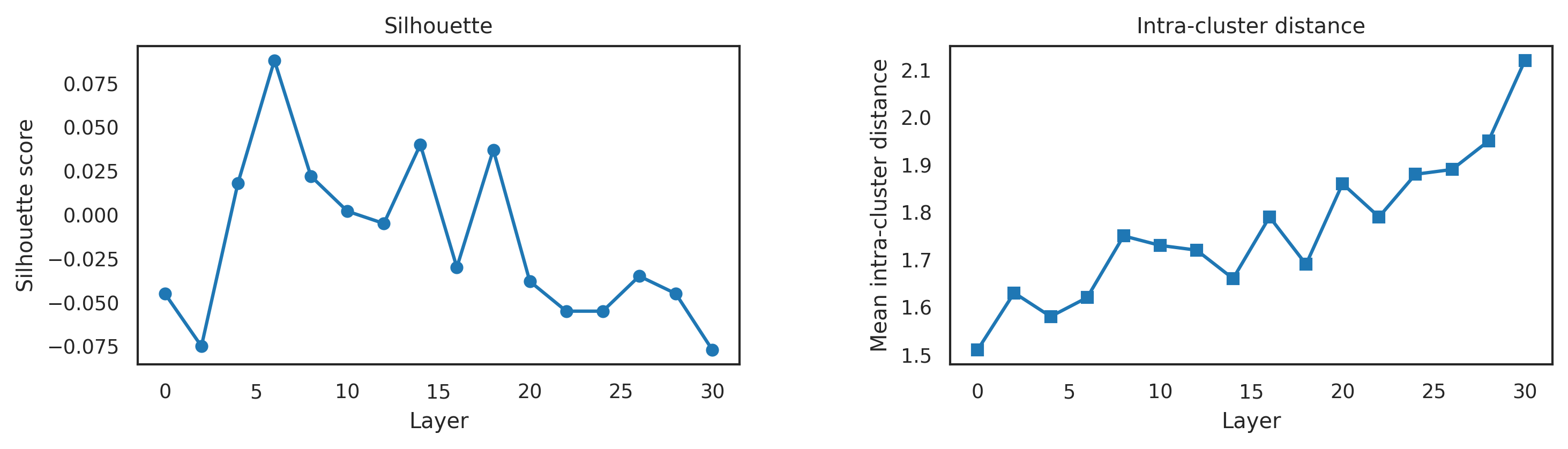}
    \caption{Layer-wise trends of UMAP-based metrics for Big Five domain labels. Left: silhouette score $S$ (higher indicates clearer separation between domains). Right: intra-cluster distance $D$ (lower indicates tighter within-domain compactness).}
    \label{fig:visualization_metrics_over_depth}
\end{figure}

Figure~\ref{fig:visualization_metrics_over_depth} shows how these metrics change across layers. The silhouette score is higher in shallow-to-mid layers and tends to decrease toward deeper layers, while intra-cluster distance generally increases with depth. This pattern is consistent with the qualitative observation that domain-wise separation is more apparent earlier and becomes more mixed in deeper layers in the UMAP visualisations.

\end{document}